\documentclass[acmsmall,screen]{acmart}

\AtBeginDocument{%
  \providecommand\BibTeX{{%
    \normalfont B\kern-0.5em{\scshape i\kern-0.25em b}\kern-0.8em\TeX}}}

\usepackage{times}
\usepackage{epsfig}
\usepackage{graphicx}
\usepackage{booktabs}

\usepackage{array}
\usepackage{amsfonts}
\usepackage{amsmath}
\usepackage{multirow}
\usepackage{pifont}
\usepackage{verbatim}
\usepackage{colortbl}
\usepackage{xcolor}
\definecolor{mygray}{gray}{.9}
\usepackage{hyperref}
\begin{document}

\title{MGRR-Net: Multi-level Graph Relational Reasoning Network for Facial Action Unit Detection}

\author{Xuri Ge}

\affiliation{%
  \institution{Unviersity of Glasgow}
  \streetaddress{School of Computing Science}
  \city{Glasgow}
  \country{UK}
}
\email{x.ge.2@research.gla.ac.uk}

\author{Joemon M. Jose}
\affiliation{%
  \institution{Unviersity of Glasgow}
\streetaddress{School of Computing Science}
  \city{Glasgow}
  \country{UK}
}
\email{joemon.jose@glasgow.ac.uk}

\author{Songpei Xu}
\affiliation{%
  \institution{Unviersity of Glasgow}
\streetaddress{School of Computing Science}
  \city{Glasgow}
  \country{UK}
}
\email{s.xu.1@research.gla.ac.uk}

\author{Xiao Liu}
\affiliation{%
  \institution{Tencent}
  \streetaddress{Online Media Business}
  \city{Beijing}
  \country{China}
}
\email{ender.liux@gmail.com}

\author{Hu Han}
\affiliation{%
  \institution{Institute of Computing Technology, Chinese Academy of Sciences and University of the Chinese Academy of Sciences}
  \city{Beijing}
  \country{China}
}
\email{hanhu@ict.ac.cn}

\renewcommand{\shortauthors}{Ge et al.}

\begin{abstract}
   The Facial Action Coding System (FACS) encodes the action units (AUs) in facial images, which has attracted extensive research attention due to its wide use in facial expression analysis.
   Many methods that perform well on automatic facial action unit (AU) detection primarily focus on modelling various AU relations between corresponding local muscle areas or mining global attention-aware facial features; however, they neglect the dynamic interactions among local-global features. We argue that encoding AU features just from one perspective may not capture the rich contextual information between regional and global face features, as well as the detailed variability across AUs, because of the diversity in expression and individual characteristics. In this paper, we propose a novel Multi-level Graph Relational Reasoning Network (termed \textit{MGRR-Net}) for facial AU detection. Each layer of MGRR-Net performs a multi-level (\textit{i.e.}, region-level, pixel-wise and channel-wise level) feature learning. On the one hand, the region-level feature learning from the local face patch features via graph neural network can encode the correlation across different AUs. On the other hand, pixel-wise and channel-wise feature learning via graph attention networks (GAT) enhance the discrimination ability of AU features by adaptively recalibrating feature responses of pixels and channels from global face features. The hierarchical fusion strategy combines features from the three levels with gated fusion cells to improve AU discriminative ability. Extensive experiments on DISFA and BP4D AU datasets show that the proposed approach achieves superior performance than the state-of-the-art methods. 
\end{abstract}

\ccsdesc[500]{Computing methodologies~Computer vision}
\ccsdesc[500]{Computing methodologies~Biometrics}
\ccsdesc[500]{Computing methodologies~Image representations}

\keywords{Facial action units, graph attention network, local-global interaction, multi-level relational reasoning}



\maketitle

\section{Introduction}
    Facial action units (AUs) are defined as a set of facial muscle movements that correspond to a displayed expression according to the Facial Action Coding System(FACS) \cite{ekman1997face}.
    As a fundamental research problem, AU detection is beneficial to facial expression analysis~\cite{yu2020facial,zhang2021short,li2021intra}, and has wide potential applications in diagnosing mental health issues \cite{rubinow1992impaired,shi2019atrial}, improving e-learning experiences \cite{niu2018automatic}, detecting deception \cite{li2016generalized}, \textit{etc.} 
    However, AU detection is challenging because of the difficulty in identifying the subtle facial changes caused by AUs and individual physiology. 
    Some earlier studies \cite{tong2008learning,li2013simultaneous} design hand-crafted features to represent different local facial regions related to AUs, according to the corresponding movements of facial muscles. 
    However, hand-crafted shallow features are not discriminative enough to represent the rich facial morphology.
    Hence, deep learning-based AU detection methods that rely on global and local facial features have been studied to enhance the feature representation of each AU.   
    \begin{figure}[t] 
	\centering
	\includegraphics[width=1.0\linewidth]{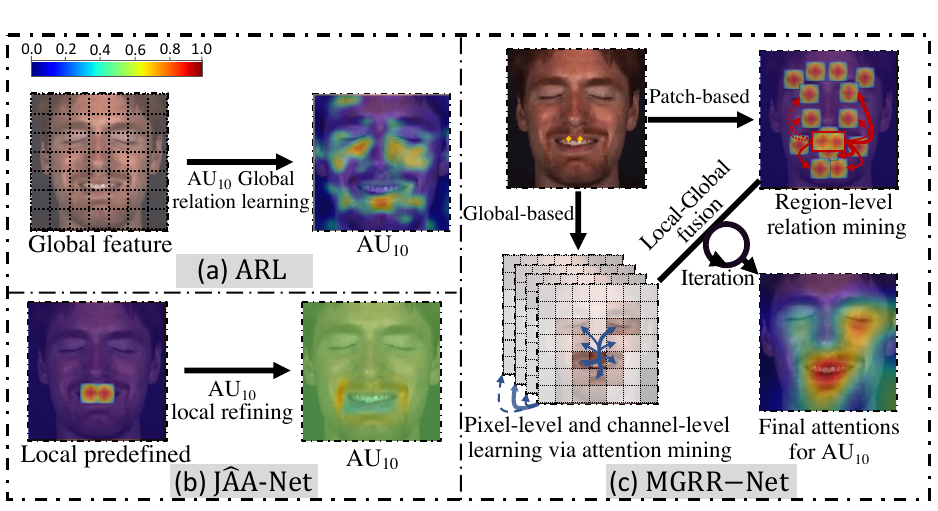}
	\vspace{-2em}
	\caption{
	Comparisons between the proposed method and two state-of-the-art methods in AU feature learning, and the corresponding visualized activation maps for AU10 (Upper Lip Raiser / Levator labii superioris). (a) ARL \cite{shao2019facial} performs global feature learning, (b) J$\rm \hat{A}$ANet \cite{shao2021jaa} learns from predefined local regions based on the landmarks, and (c) multi-level feature learning from both local regions and global face regions (best viewed in color).
    }
	\label{fig:frams}
	\vspace{-0.5em}
\end{figure} 

    Several recent works \cite{liu2014feature,shao2019facial,niu2019local,sankaran2020domain} aim to enhance the corresponding AU feature representation by combining the affected features in a deep global face feature map. 
    For instance, LP-Net \cite{niu2019local} using an LSTM model \cite{hochreiter1997long} combines the patch features from  grids of equal partition made by a global Convolutional Neural Network (CNN). 
    ARL \cite{shao2019facial} directly learns spatial attention from the global CNN features of independent AU branches, as shown in  Fig. \ref{fig:frams} (a). 
    And \cite{luo2022learning} separately represented AU features directly from a shared full-face feature via multiple independent fully connected layers to model the relationships among all AUs in a graph. 
    However, these methods suffered from the challenges of accurate localization of muscle areas corresponding to AUs, leading to potential interference from some irrelevant regions. 
    In the past, such issues were addressed by extracting AU-related features from regions of interest (ROIs) centered around the associated facial landmarks \cite{zhao2016joint,shao2018deep,shao2021jaa}, which provide more precise muscle locations for AUs and lead to a better AU detection performance. 
    For example, JAA \cite{shao2018deep} and J$\rm \hat{A}$ANet \cite{shao2021jaa} propose attention-based deep models to adaptively select the highly-contributing neighboring pixels of initially predefined muscle region for joint AU detection and face alignment, as shown in the Fig. \ref{fig:frams} (b).
    However, the above local attention-based methods emphasize learning the appearance representation of each facial region based on detected landmarks while ignoring some intrinsic dependencies between different facial muscles. 
    For example, AU2 (``Outer Brow Raiser") and AU7 (``Lid Tightener") will be activated simultaneously when scaring and AU6 (``Cheek Raiser") and AU12 (``Lip Corner Puller"), usually simultaneously in a smiling face. 
    To this end, some methods \cite{zhang2018classifier,niu2019multi,liu2020relation,cui2020knowledge} try to utilize prior knowledge of AU correlation by defining a fixed graph that represents the statistical AU correlations. 
    For instance, \cite{liu2020relation} constructs a predefined graph for each face based on the AU co-occurrences to explicitly model the relationships between AU regions and enhance their semantic representations.  
    However, it is  difficult to  effectively capture  the dynamic relationships between AUs and the distinction of related AUs by a single predefined graph due to the complexity of AU activation and diversity across different subjects. 
    Recent works \cite{song2021uncertain,song2021dynamic,song2021hybrid} make an attempt to exploit an adaptive graph to model the uncertainty relationship between AUs.
    For instance, \cite{song2021dynamic} emphasises the learning of important local facial regions based on probabilistic graph and obtain  better facial appearance features by emphasizing important local facial regions via Long Short-Term Memory (LSTM) \cite{graves2005framewise}. 
    However, these approaches still enhance the semantic AU representations from the perspective of better regional feature representation, neglecting the modelling of the distinctive local and global features of each AU.

    The key issue of facial AU detection lies in obtaining a better facial appearance representation by improving the feature discriminative ability of local AUs and global features from the whole face. 
    On the one hand, region-level dynamic AU relevance mining based on facial landmarks accurately detects the corresponding muscles and flexibly models the relevance among muscle regions.  
    It is different from the existing methods focusing on extracting features for a single AU region  \cite{shao2018deep, shao2021jaa} or a predefined fixed graph representing prior knowledge \cite{li2019semantic}. 
    Although there have been many methods \cite{liu2020relation,song2021dynamic,song2021uncertain,song2021hybrid} on modelling relationships between AU regions, this issue still needs to be addressed effectively. 
    On the other hand, due to the differences in expressions, postures and individuals, fully learning the responses of the target AU in the global face can better capture the contextual differences between different AUs and complement more semantic details from the global face. 
    For instance, \cite{shao2018deep,shao2021jaa} simply concatenated the global features extracted from the whole face via CNNs with all local AU features for input into the final classifier. 
    However, it is difficult for all these methods to learn the sensitivity of the target AU within the global face and supplement enough semantic details from the global face representation in terms of different expressions, postures and individuals. 
    To the best of our knowledge, how to better respond globally to each AU remains unexploited in existing works \cite{liu2020relation,li2019semantic,shao2021jaa,luo2022learning}.

    Motivated by the above insights, we propose a novel technique for facial AU detection called  MGRR-Net. 
    Our main innovations lie in three aspects, as shown in  Fig. \ref{fig:frams} (c). 
    Firstly, we introduce a dynamic graph to model and reason the relationship between a target AU and other AUs. 
    The region-level AU features (as nodes) can accurately locate the corresponding muscles. 
    Secondly, we supplement each AU with different levels (channel- and pixel-level) of attention-aware details from global features, which greatly improves the distinction between AUs.
    Finally, we iteratively refine the AU features of the proposed multi-level local-global relational reasoning layer, which makes them more robust and more interpretable. 
    Different from the existing GNN-based approaches \cite{li2019semantic,niu2019multi,song2021hybrid, song2021uncertain, liu2020relation,luo2022learning} that utilize complex GCNs \cite{kipf2016semi} to enhance the distinguishability of AUs by constructing AU relationships, however, we supplement each AU with different perspectives (channel- and pixel-level) of attention-aware details from global features, making it possible to achieve the same purpose in a basic GNN and solve a certain over-smoothing issue. 
    In particular, we extract the global features by multi-layer CNNs and precise AU region features based on the detected facial landmarks, which serve as the inputs of each multi-level relational reasoning layer. 
    A simple region-level AU graph is constructed to represent the relationships by the adjacency matrix (as edges) among AU regions (as nodes), initialized by prior knowledge and iteratively updated. 
    We propose a method to learn channel- and pixel-wise semantic relations for different AUs at the same time by processing them in two separate efficient and effective multi-head graph attention networks (MH-GATs) \cite{velivckovic2018graph}. Through this, we model the complementary channel- and pixel-level global details. 
    After these local and global relation-oriented modules, a hierarchical gated fusion strategy helps to select more useful information for the final AU representation in terms of different individuals. 
    
    The contributions of this work are as follows: 
  
    \begin{itemize}
    \item We propose a novel end-to-end iterative reasoning and training scheme for facial AU detection, which leverages the complementary multi-level local-global feature relationships to improve the robustness and discrimination for AU detection;
    \item We construct a region-level AU graph with the prior knowledge initialization and dynamically reason the correlated relationship of individual AUs, thereby improving the robustness of AU detection;
    \item We propose a GAT-based model to improve the discrimination of each local AU patch by supplementing multiple levels of global features; 
    \item The proposed MGRR-Net outperforms the state-of-the-art approaches for AU detection on two widely used benchmarks, \textit{i.e.,} BP4D and DISFA, without any external data or pre-trained models.
    \end{itemize}

\section{Related Work}
    \subsection{Facial AU Detection} 
    Automatic AU detection has been studied for decades, and several methods \cite{zhao2016joint,ma2019r,shao2019facial,niu2019local,zhang2021facial,li2021micro,shao2021jaa} have been proposed.  
    Some works \cite{liu2014feature,shao2019facial,niu2019local,sankaran2020domain,li2021micro} predicted the activation state of each AU by directly extracting global face features via CNNs.
    For instance, \cite{shao2019facial,li2021micro} proposed  sequential or parallel channel and spatial attention learning mechanisms to explore the attention-aware global representation of each face. 
    While progress has indeed been achieved through the utilization of global representations, the advancement remains constrained by the rudimentary nature of the coarse-grained features.
    Most existing approaches for facial AU detection use feature learning from local patches \cite{zhao2016joint,li2017action,shao2018deep, ma2019r,niu2019local,shao2021jaa}. 
       However, there is a need to pre-define the patch location first in some early works \cite{taheri2014structure, zhao2015joint}. 
    For instance, \cite{jaiswal2016deep} proposed to use domain knowledge and facial geometry to pre-select a relevant image region (as a patch) for a particular AU and feed it to a convolutional and bi-directional Long Short-Term Memory (LSTM) \cite{graves2005framewise} neural network. 
    \cite{shao2018deep} proposed an end-to-end deep learning framework for joint AU detection and face alignment, which used the detected landmarks to locate specific AU regions. 
        However, all the above methods focused only on independent regions without considering the correlations among different AU areas to reinforce and diversify each other. 
      Recent works focus on capturing the relations among AUs for local feature enhancement, which can improve robustness compared to single-patch features or global face features. 
    \cite{li2019semantic} incorporated the AU knowledge graph as extra guidance for enhancing facial region representation. 
    \cite{liu2020relation} applied the spectral perspective of graph convolutional network (GCN) for AU relation modelling, which also needed an additional AU correlation reference extracted from EAC-Net \cite{li2018eac}. 
    However, these methods need  prior knowledge of co-occurrence probability in different datasets to construct the fixed relation matrix instead of dynamically updating for different expressions and individuals. 
    \cite{ge2021local} proposed a complex skip-BiLSTM to mine the potential mutual assistance and exclusion relationship between AU branches and simple complementary global information. 
    \cite{song2021hybrid} proposed a performance-driven Monte Carlo Markov Chain to generate graphs from the global face, which, however, also captures some irrelevant regions affecting the performance. 
    Moreover, these approaches usually ignored or simply fused the local and global information for each AU without considering the importance (important and non-important) of features. 
    Recently, \cite{luo2022learning} learned a unique AU graph to explicitly describe the relationship between AUs, where each AU is simply represented from the same full face representation via a fully connected layer and a global average pooling. Although this method explores the global face features to some extent, it relies on strong global feature extraction benchmarks and lacks accurate localization of local muscle areas and  discriminable feature representation via local-global interaction.

    \subsection{Graph Neural Network}
     Integrating graphs with deep neural networks have recently been an emerging topic in deep learning research. 
    GCNs have been widely used in many applications such as human action recognition \cite{yan2018spatial}, emotion recognition \cite{song2018eeg}, social relationship understanding \cite{wang2018deep} and object parsing \cite{liang2016semantic}. 
    \cite{li2019semantic} proposed to apply a gated graph neural network (GGNN) with the guidance of AU knowledge-graph on facial AU detection. 
    \cite{niu2019multi} embedded the relations among AUs through a predefined GCN to enhance the local semantic representation. 
    However, these AU detection methods require a fixed predefined graph from different datasets when applying GGNN or GCN. 
    \cite{song2021hybrid,song2021uncertain} applied an adaptive graph to model the relationships between AUs based on global features, ignoring local-global interactions.  
    Recently, a novel graph attention network with multi-head (MH-GAT) leverages masked self-attentional layers to operate on graph-structured data, which shows high computational efficiency.

    As far as we know, there has been no work attempting to obtain  better feature representation by multiple interactions between local AU regions and the global face, which we believe is an important cue to boost facial AU detection performance with more fine-grained information and higher diversity of expressions. 
    To this end, our proposed MGRR-Net automatically models the relevance among the facial AU regions by a dynamic matrix as a graph and supplements each AU patch with multiple levels of global features to improve the variability. 
    Multiple layers of iterative refinement significantly improve the AU discrimination ability. 
    Our MGRR-Net has wide potential applications in diagnosing mental health issues \cite{rubinow1992impaired,shi2019atrial}, improving e-learning experiences \cite{niu2018automatic}, detecting deception \cite{li2016generalized}, \textit{etc.} 
    For example, in our future work, we will apply MGRR-Net to automatically estimate facial palsy severity for patients, such as \cite{ge2023algrnet}. This will be helpful for the diagnosis and treatment of people who have facial palsy across the world.

 \section{Approach}
\begin{figure*}[t] 
	\centering
	\includegraphics[width=1\linewidth]{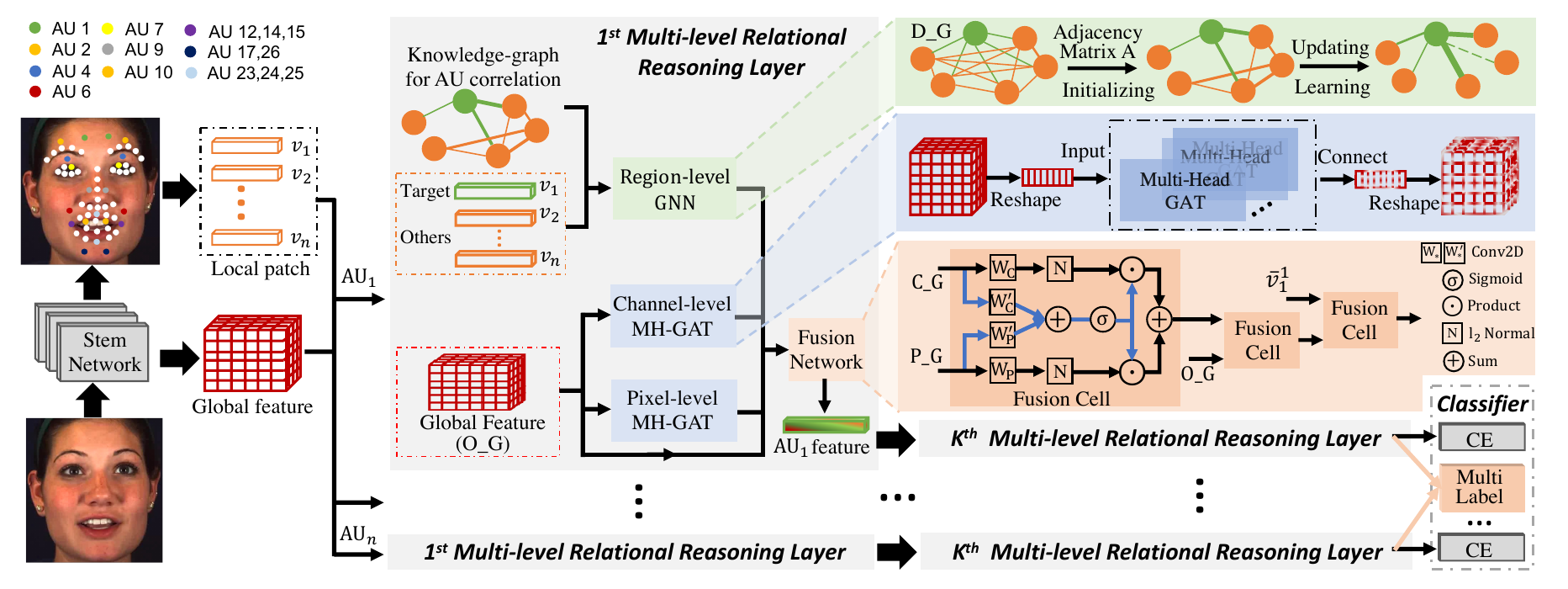}
	\caption{The overall architecture of the proposed MGRR-Net for facial AU detection. 
	Given one face image, the region-level features of local AU patches are extracted based on the detected landmarks from an efficient landmark localization network. The original global feature is extracted from the same shared stem network. Then the region-level GNN  initialized with prior knowledge is applied to encode the correlation between different AU patches. Two separate MH-GATs are adopted to get two levels of global attention-aware features to supplement each AU. Finally, multiple levels of local-global features are fused by a hierarchical gated fusion strategy and refined by multiple iterations (best viewed in color). 
	}
	\label{fig:fig_overall}
\end{figure*}

    As shown in Fig. \ref{fig:fig_overall}, the proposed approach consists of two core modules in each relational reasoning layer, \textit{i.e}., region-level local feature learning with relational modelling, and global feature learning with channel- and pixel-level attention. 
    A hierarchical gated fusion network is designed to combine multi-level local and global features as the new target AU feature. 
    Finally, after multiple layers of iterative refinement and updating, the AU features are fed into a multi-branch classification network for AU detection.
    For clarity, the main notations and their definitions throughout the paper are shown in Table \ref{tab:notations}.

    \subsection{Global and Local Features Extraction} \label{Overview of MGRR-Net}
     Given a face image $I$, we adapt a stem network from the widely used multi-branch network \cite{shao2018deep} to extract the original global feature $\rm O\_G$ and further obtain the AU regions based on the detected landmarks. 
     Different from the \cite{li2018eac}, our stem network contains a face alignment module for automatic face landmark detection, facilitating end-to-end training of our method. 
        All branches share the stem network to reduce training costs and the complexity of network training.
        In particular, a hierarchical and multi-scale region learning module in the stem network extracts features from each local patch with different scales, thus obtaining multi-scale representations.
        A series of landmarks $S = \{s_1,s_2,...,s_m\}$ with length $m$ are detected by an efficient face alignment module similar to \cite{shao2021jaa}, including three convolutional blocks connected to a max-pooling layer.
        According to the detected landmarks, local patches are calculated, and their features $V=\{v_1,v_2,...,v_n\}$ are learned via the stem network, where $n$ is the number of selected AU patches. 
        For simplicity, we do not repeat the detailed structure of the stem network here.
\begin{table}[t]
\scriptsize
\begin{center}
\fontsize{9.5}{13}\selectfont
\renewcommand\tabcolsep{4.0pt}
\caption{Main notations and their definitions.} \label{tab:notations}
\begin{tabular}{l|l}
\hline

\hline
Notation & Definition \\ \hline \hline
$I$         &  a facial image         \\
$S$         &  a set of detected landmarks         \\
$\rm{O\_G}$ &  the original global feature   \\
$m$         &  the number of detected landmarks \\
$V$         &  a set of calculated patch features \\
$v_i$       &  the feature of $i$-th patch  \\
$n$         &  the number of calculated patches corresponding to AUs \\
$\rm{D\_G}$  &  a fully-connected graph for AU relationship construction  \\
$A$         & a learnable adjacency matrix  \\
$a_i$       & the activation status of the $i$-th AU \\
$P_{ij}$    & the coefficient between $i$-th and $j$-th AU \\
$P, C$     & a set of pixel- and channel-level features  \\
$\rm{P\_G}$       &  the pixel-level attention-aware global feature   \\
$\rm{C\_G}$       &  the channel-level attention-aware global feature  \\
$L$         & the number of parallel attention layers  \\
$K$         &  the number of relational reasoning layers        \\
$\Bar{v}_i^{k}$  &  the feature of $i$-th AU patch after $k$-th reasoning layer        \\
$\rm{GFC}$       & a gated fusion cell \\
$(x_i,y_i)$ &  the ground-truth coordinate of the $i$-th facial landmark\\
$(\hat{x}_i,\hat{y}_i)$ & the predicted coordinate of the $i$-th facial landmark \\
$d_o$      & the ground-truth inter-ocular distance\\
$p_i$     & the ground-truth occurrence probability of $i$-th AU \\
$\hat{p}_i$ & the predicted occurrence probability of $i$-th AU \\
\hline
         
\hline
\end{tabular} 
\end{center}
\end{table}           
 \subsection{Multi-level Relational Reasoning Layer} \label{MRRL}
        After we get the original global feature $\rm O\_G$ for a face and the local region features $V=\{v_1,v_2,...,v_n\}$ for AUs, a multi-layer multi-level relational reasoning model is introduced to automatically explore the relationship of individual local facial regions and supply two levels of global information. 
        Fig. \ref{fig:fig_overall} shows the detailed structure of the $1^{st}$ multi-level relational reasoning layer.

        \subsubsection{Region-level Local Feature Relational Modeling}
         Different from the predefined fixed AU relationship graph in \cite{li2019semantic}, we construct a fully-connected graph $\rm D\_G$ for all AUs, where the region-level features $V=\{v_1,v_2,...,v_n\}$ constitute the nodes, and a learnable adjacency matrix $A$ constitutes the edges at each layer to represent the possibility of AU co-occurrence (co-activated or non-activated). In this scheme, the AUs with no co-occurrence or low co-occurrence relationship in the training set will not be completely ignored, like \cite{li2019semantic,niu2019multi,liu2020relation}.
         During the training process, we utilize prior knowledge to initialize $A$ to assist and constrain model learning. 
         Specifically, the dynamic graph $\rm D\_G$ comprises nodes (the local region features $V=\{v_1,v_2,...,v_n\}$) and edges (the relationship matrix $A$ among AUs). 
         Following \cite{niu2019multi}, we calculate the relationship coefficients between AUs from datasets to initialize the adjacency matrix $A$ (Fig. \ref{fig:visual_02} (a) shows the predefined AU correlation on BP4D). 
          The statistical prior knowledge serves as the initial relationship, allowing suppression of the edges with low correlation and speeding up the relationship learning. 
         The relationship coefficient $A_{ij}$ between the $i$-th and $j$-th AU can be formulated as:
        \begin{equation} \label{p1}
            P_{ij} = \frac{1}{2} (P(a_i=1 | a_j=1) + P(a_i=0 | a_j=0)),
        \end{equation}     
        \begin{equation} \label{p2}
            A_{ij} = |(P_{ij} - 0.5) * 2|
        \end{equation}      
        where $a_i$=1 denotes $i$-th AU is activated and 0 otherwise, $|\cdot|$ means absolute value function. 
        From Eq.\ref{p1} and Eq.\ref{p2}, $P (a_i$=$1|a_j$=$1)$=$0.5$ means that when $j$-th AU is activated, the probability of occurrence is equal to the no occurrence for $i$-th AU. 
        It indicates that the activation of $j$-th AU could not provide useful information for the $i$-th AU, and therefore no edge is connected.

        \subsubsection{Attention-aware Global Features Learning}
        We argue that complementary  global feature can improve the discrimination between AUs, which also alleviates the over-smoothing issue in graph neural networks for local relationship modelling. 
        To this end, we employ two separate high-efficiency GAT models \cite{velivckovic2018graph} to perform channel- and pixel-level attention-aware global features from original deep visual features in order to handle expression and subject diversities. 
        Specifically, we reshape the original global feature $\rm O\_G \in \mathbb{R}^{(c,w,h)}$ into a set of channel-level features $\{C_1,...,C_c\},C_i\in\mathbb{R}^{w*h}$. 
        Similarly, by reshaping pixel dimensions and keeping channel dimension of $\rm O\_G$ from a convolution layer to reduce the parameters, we get a set of pixel-level features $\{P_1,...,P_{w'*h'}\},P_i\in\mathbb{R}^{c}$. 
        The attention coefficient $\alpha_{ij}$ between channel- or pixel-level features is calculated in GAT, which can be formulated as (Here we take the process of channel-level attention-aware features as an example.):    
        \begin{equation}
            \alpha_{ij}=  \frac{ {\rm exp}(U_q C_i(U_k C_j)^T / \sqrt{D})}{{\rm \sum_{o\in \Omega_i} exp}(U_q C_i(U_o C_o)^T / \sqrt{D})},
        \end{equation}             
        where $U_q,U_k,U_o$ are the parameters of mapping from $w*h$ to $D$ and $\Omega_i$ denotes neighborhoods of $C_i$. $\sqrt{D}$ acts as a normalization factor. 
        Following \cite{vaswani2017attention,velivckovic2018graph}, we also employ multi-head dot product by $L$ parallel attention layers to speed up the calculation efficiency. The overall working flow is formulated as: 
        \begin{equation}
            \begin{aligned}
            \Bar{C}_i = {\rm ReLU}({\rm \sum\nolimits_{o\in \Omega_i}} U_c||_l^L(\alpha_{io}^l * C_i)), \\
            \alpha_{ij}^l=  \frac{ {\rm exp}(U'_q C_i(U'_k C_j)^T / \sqrt{d})}{{\rm \sum\nolimits_{o\in \Omega_i} exp}(U'_q C_i(U'_o C_o)^T / \sqrt{d})},
            \end{aligned}
        \end{equation}       
        where $U_c$ is the mapping parameter, $U'_q,U'_k,U'_o$ map the feature dimension to $1/L$ of the original, $||$ means concatenation, and $d$ equals $D/L$. 
        Finally, the new channel-level attention-aware global feature $\rm C\_G$=$\{\Bar{C}_i\}$ is reshaped to the same domination with $\rm O\_G$. 
        With the same process on pixel-level features $\{P_1,...,P_{w'*h'}\}$, we can get the final pixel-level attention-aware global features $\rm P\_G$ after a deconvolution layer behind of a GAT with multi-head (MH-GAT).

    \subsubsection{Hierarchical Fusion and Iteration}
           We iteratively refine the $i$-th target AU feature of the proposed multi-level relational reasoning layer $K$ times, which obtains other correlated local, and regional information and provides rich global details in each layer. 
        The process can be formulated as:
        \begin{equation}
            \Bar{v}_i^{k} = W_i^k v_i^k + \sum\nolimits_i^n(A^k_{ij} W_j^k v_j),
        \end{equation}      
        where $W^k$ is the mapping parameter and $A^k_{ij}$ means the learnable correlation coefficient between AU$_i$ and AU$_j$ at $k$-th layer. 
        We then use a hierarchical fusion strategy by a gated fusion cell (GFC) to complement the global multi-level information for each updated AU feature at $k$-th layer as follows:
        \begin{equation}
            \Bar{v}_i^{k+1} = {\rm GFC}(\Bar{v}_i^{k},{\rm GFC}({\rm O\_G}^k, {\rm GFC}({\rm C\_G}^k,{\rm P\_G}^k))),
        \end{equation}         
        We define the operation of ${\rm GFC}$ as follows:
        \begin{equation}
        \begin{aligned}
            {\rm GFC}&({\rm C\_G}^k, {\rm P\_G}^k)   = \beta \odot \lVert W_C^k {\rm C\_G}^k\lVert_2 + (1-\beta) \odot \lVert W_P^k {\rm P\_G}^k\lVert_2,
        \end{aligned}
        \end{equation}  
        \begin{equation}
            \beta = {\rm {\sigma}}({W_C^{k'}} {\rm C\_G}^k+{{W^{k'}_P}} {\rm P\_G}^k),
        \end{equation}         
        where $\sigma$ is the sigmoid function, and $\lVert \cdot \lVert$ denotes the $l_2$-normalization. $ {W^{k'}_*}$ and $ {W^k_*}$ denote the Conv2D operation. 
        
        \subsection{Joint Learning}
        A multi-label binary classifier is used to classify the AU activation state, which adopts a weighted multi-label cross-entropy loss function (denoted as CE in Fig. \ref{fig:fig_overall}) as follows,
        \begin{equation} \label{wCE}
             \mathcal{L}_{au} = - \frac{1}{n} \sum_{i=1}^{n} w_i [p_i {\rm log} \hat{p_i} + (1-p_i) {\rm log} (1-\hat{p}_i)],
        \end{equation} 
        where $p_i$ and $\hat{p}_i$ denote the ground-truth and predicted occurrence probability of the $i$-th AU, respectively; $w_i$ is the data balance weights used in \cite{shao2018deep}. 
        Furthermore, we also minimize the loss of AU category classification $\mathcal{L}_{int}$ by integrating all AUs information, including the refined AU features and the face alignment features, which is similar to the processing of $\mathcal{L}_{au}$.

        We jointly integrate  face alignment and facial AU recognition into an end-to-end learning model.
        The face alignment loss is defined as:     
        \begin{equation}
            \label{eq:Lalign} \mathcal{L}_{align} = \frac{1}{2d_o^2} \sum_{i=1}^{m} [(x_i-\hat{x_i})^2+(y_i-\hat{y_i})^2],
        \end{equation}
        where $(x_i,y_i)$ and $(\hat{x}_i,\hat{y}_i)$ denote the ground-truth coordinate and corresponding predicted coordinate of the $i$-th facial landmark, and $d_o$ is the ground-truth inter-ocular distance for normalization \cite{shao2021jaa}. 
        Finally, the joint loss of our MGRR-Net is defined as:
        \begin{equation}
            \label{eq:Lall} \mathcal{L} = (\mathcal{L}_{au} + \mathcal{L}_{int})+ \lambda \mathcal{L}_{align}.
        \end{equation}         
        where $\lambda$ is a tuning parameter for balancing.

\section{Experiments}
 In this section, we conduct extensive experiments to evaluate the proposed MGRR-Net. 
    Especially the dataset and training strategy are first introduced. 
    Then, MGRR-Net is compared with state-of-the-art FAU detection approaches quantitatively.
    Finally, we qualitatively analyze the results in detail.
    
    \subsection{Dataset}
        We provide evaluations on the popular BP4D \cite{zhang2014bp4d} and DISFA \cite{mavadati2013disfa} datasets.
        
        \textbf{BP4D} is a spontaneous facial AU database containing $328$ facial videos from 41 participants (23 females and 18 males) who were involved in 8 sessions. 
        Similar to \cite{li2017action,shao2019facial,shao2021jaa}, we consider 12 AUs and 140K valid frames with labels. 
        
        \textbf{DISFA} consists of 27 participants (12 females and 15 males). Each participant has a video of $4,845$ frames. 
        We limited the number of AUs to 8, similar to \cite{li2017action,shao2021jaa}. 
        Following \cite{shao2018deep,shao2021jaa}, frames in DISFA with AU intensity labels higher than two are considered positive samples. 
      Compared to BP4D, the experimental protocol and lighting conditions deliver DISFA to be a more challenging dataset.
      
        During training, each frame of BP4D and DISFA is annotated with 49 landmarks detected and calculated by SDM \cite{xiong2013supervised}. 
        Following the experiment setting of \cite{shao2018deep,shao2021jaa}, we evaluated the model using the 3-fold subject-exclusive cross-validation protocol.

    \subsection{Training Strategy}
        Our model is trained on a single NVIDIA RTX 2080Ti with 11 GB memory.
        The whole network is trained with the default initializer of PyTorch \cite{paszke2019pytorch} with the SGD solver, a Nesterov momentum of 0.9 and a weight decay of 0.0005. 
        The learning rate is set to 0.01 initially, with a decay rate of 0.5 every two epochs.
        The maximum epoch number is set to 15. 
        During the training process, aligned faces are randomly cropped into $176 \times 176$ and horizontally flipped. 
        Regarding the face alignment network and stem network, we set the value of the general parameters to be the same with \cite{shao2021jaa}. 
        The iteration layer number $K$ is set to 2 except otherwise noted. 
        The dimensionality of $O\_G$ is $(64,44,44)$ and $D$ is 1024.
        We employ $L$=8 parallel attention layers in GATs. 
           In our paper, all the mapping Conv2D operations used $1 \times 1$ convolutional filters with a stride one and a padding 1. We use a $3 \times 3$ Conv2D operation with a stride two and padding one before learning the channel-level feature to reduce the parameters.
        $\lambda$ is empirically set to 0.5 for the joint optimisation of face alignment and facial AU detection on two benchmarks. 
        Following the settings in \cite{zhao2016deep, li2018eac, shao2021jaa}, our MGRR-Net initializes the parameters of the well-trained model trained on BP4D when training on DISFA. This initialization greatly alleviates the poor performance issue on DISFA due to data volume and AU category imbalance. 
        Compared to $\rm J\hat{A}A$-$\rm Net$ \cite{shao2021jaa}, which takes 26.6ms per image to do a forward pass, our model takes just 16.5ms using an RTX 2080Ti GPU. This is due to the multi-head operation of the effective MH-GATs and the optimization of the model, which significantly reduces forward pass time. 
        The training time is approximately 1.5 hours per epoch.
        In addition, we average the predicted probability of the local information and the integrated information as the final predicted activation probability for each AU rather than simply using the integrated information of all the AUs. 
\begin{table*}[t]
\centering
\fontsize{9.5}{12.5}\selectfont
\renewcommand\tabcolsep{8pt}
\caption{Comparisons of AU recognition for 8 AUs on DISFA  in terms of F1-frame score (in \%).  CLP$^{\dag}$ is a semi-supervised method. * means the method employed a pre-trained model on the additional dataset, such as ImageNet \cite{deng2009imagenet} and VGGFace2 \cite{cao2018vggface2}, \textit{etc.}} 
\label{tab:DISFA}
 
\begin{tabular}{cccccccccc}
\hline

\hline
\multirow{2}{*}{Method} & \multicolumn{8}{c}{AU Index}                          & \multirow{2}{*}{\textbf{Avg.}} \\ \cline{2-9}
                        & 1 & 2 & 4 & 6 & 9 & 12 & 25 & 26 &                       \\ \hline
DSIN \cite{corneanu2018deep} & 42.4 & 39.0 & 68.4 & 28.6 & 46.8 & 70.8 & 90.4 & 42.2 & 53.6 \\
JAA \cite{shao2018deep}  & 43.7 & 46.2 & 56.0 & 41.4 & 44.7 & 69.6 & 88.3 & 58.4   & 56.0        \\
LP-Net \cite{niu2019local}   & 29.9 & 24.7 & {72.7} &  46.8 & {49.6} & 72.9 & 93.8 & 65.0   & 56.9                   \\
ARL \cite{shao2019facial}    & 43.9 & 42.1 & 63.6 & 41.8 & 40.0 & \textbf{76.2} & \textbf{95.2} & 66.8   & 58.7                  \\
SRERL \cite{li2019semantic}  & 45.7 & 47.8 & 59.6 & \underline{47.1} & 45.6 & 73.5 & 84.3 & 43.6 & 55.9 \\
J$\rm \hat{A}$ANet \cite{shao2021jaa}        & \textbf{62.4} & \underline{60.7} & 67.1 & 41.1 & 45.1 & 73.5 & 90.9 & {67.4}   & {63.5}            \\ 
JAA-DGCN \cite{jia2023novel}  &  \underline{61.8}  & 51.7 &  64.5  & 46.0  & \underline{54.2}  & 63.6  & 85.5  & 69.4 &  62.0 \\
CLP$^{\dag}$ \cite{li2023contrastive}  & 42.4  & 38.7  & 63.5  &  59.7  &  38.9  &  73.0  &  85.0  &  58.1  &  57.4 \\
MMA-Net \cite{shang2023mma}  &  63.8 &  54.8 &  \underline{73.6} &  39.2 &  \textbf{61.5} &  73.1 &  92.3 &  \underline{70.5} &  \underline{66.0} \\
\rowcolor{gray!20} \textbf{MGRR-Net}            &  {61.3}  &  \textbf{62.9}  &  \textbf{75.8}  &  \textbf{48.7}  &  {53.8}  & \underline{75.5} &  \underline{94.3}  &  \textbf{73.1}  &  \textbf{68.2}                      \\ \hline
 
 UGN-B* \cite{song2021uncertain}  & 43.3 & 48.1 & 63.4 & 49.5 & 48.2 & 72.9 & 90.8 & 59.0 & 60.0 \\
 HMP-PS* \cite{song2021hybrid} & 21.8 & 48.5 & 53.6 & 56.0 & 58.7 & 57.4 &  55.9 & 56.9 & 61.0 \\
 DML* \cite{wang2021dual} & 62.9 & 65.8 & 71.3 & 51.4 & 45.9 & 76.0 & 92.1 & 50.2 & 64.4 \\
 PIAP* \cite{tang2021piap} & 50.2 & 51.8 & 71.9 & 50.6 & 54.5 & 79.7 & 94.1 & 57.2 & 63.8 \\
 TransAU* \cite{jacob2021facial} & 46.1 & 48.6 & 72.8 & 56.7 & 50.0 & 72.1 & 90.8 & 55.4 & 61.5  \\ 
 Bio-AU* \cite{cui2023biomechanics}  & 41.5 & 44.9 & 60.3 & 51.5 & 50.3 & 70.4 & 91.3 & 55.3 & 58.2 \\
 \rowcolor{gray!20}  \textbf{MGRR-Net}   &  \underline{61.3}  &  \underline{62.9}  &  \textbf{75.8}  &  48.7  &  {53.8}  &  {75.5}  &  \underline{94.3}  &  \textbf{73.1}  &  \textbf{68.2}  \\ \hline

\hline
\end{tabular}
 
\end{table*}
\begin{table}[t]
\centering
\setlength\tabcolsep{5pt}
\fontsize{9.5}{12.5}\selectfont
\caption{Comparisons of AU recognition for 8 AUs on DISFA in terms of Accuracy and AUC (in \%). * means the method employed pretrained model on additional dataset.}
\label{tab:DISFA_ACC_AUC}
\begin{tabular}{c|ccccc >{\columncolor{gray!20}} c|cccc >{\columncolor{gray!20}} c}
\hline

\hline
\multirow{2}{*}{AU} & \multicolumn{6}{c|}{Accuracy}       & \multicolumn{5}{c}{AUC} \\ \cline{2-12} 
                    & \rotatebox{90}{JAA \cite{shao2018deep}} & \rotatebox{90}{ARL \cite{shao2019facial}} & \rotatebox{90}{J$\rm \hat{A}$ANet } & \rotatebox{90}{MMA-Net \cite{shang2023mma}} & \rotatebox{90}{UGN-B* \cite{song2021uncertain}} & \rotatebox{90}{\textbf{MGRR-Net}} & \rotatebox{90}{DRML \cite{zhao2016deep}}   & \rotatebox{90}{SRERL \cite{li2019semantic}} & \rotatebox{90}{DML* \cite{wang2021dual}}  & \rotatebox{90}{DAR-GCN \cite{jia2022data}} & \rotatebox{90}{\textbf{MGRR-Net}}  \\ \hline
1                   & 93.4    & 92.1    & \textbf{97.0} & \underline{96.8} & 95.1       &  \underline{96.8}      & 53.3       & {76.2}   & \textbf{90.5}   & 84.5  &  \underline{89.5}       \\
2                   & 96.1    & 92.7    & \underline{97.3} &  96.5 & 93.2       &  \textbf{97.4}      & 53.2       & 80.9   & \underline{92.7}  & 92.5 &  \textbf{93.0}       \\
4                   & 86.9    & \underline{88.5}    & 88.0  & 91.6 & 88.5       &  \textbf{92.7}      & 60.0       & {79.1}   & \textbf{93.8}   & 72.2 &  \underline{93.6}       \\
6                   & 91.4    & 91.6    & {92.1}  & 91.5 & \textbf{93.2}       &  \underline{92.1}      & 54.9       & 80.4   & \underline{90.3} & 48.3 &  \textbf{91.1}       \\
9                   & 95.8    & 95.9    & 95.6  & 96.5 & \underline{96.8}       &  \textbf{96.9}      & 51.5       & 76.5   & \underline{84.4}   & 78.3 &  \textbf{91.9}       \\
12                  & 91.2    & \textbf{93.9}    & 92.3   &  92.3 & 93.4       &  \underline{93.4}      & 54.6       & 87.9  & \underline{95.7}    & 37.8 &  \textbf{95.9}       \\
25                  & 93.4    & \textbf{97.3}    & 94.9  & 95.5  & 94.8       &  \underline{96.8}      & 45.6       & 90.9   & \underline{98.2}   & 50.3 &  \textbf{99.0}       \\
26                  & 93.2    & 94.3    & {94.8} & \underline{95.0} & 93.8       &  \textbf{95.6}      & 45.3       & 73.4   & \underline{87.4}   & 74.3 &  \textbf{94.4}       \\ \hline
\textbf{Avg.}                & 92.7    & 93.3    & {94.0} & \underline{94.5} & 93.4       &  \textbf{95.2}      & 52.3       & 80.7   & \underline{91.6}   & 67.3 & \textbf{93.6}      \\ \hline

\hline
\end{tabular}
\end{table}

\begin{table*}[t]
\centering
\caption{Comparisons with state-of-the-art methods for 12 AUs on BP4D in terms of F1-frame (in \%). * means the method employed pretrained model on additional dataset.} 
\label{tab:BP4D_f1}
\setlength\tabcolsep{2.8pt}
\fontsize{9.5}{12.5}\selectfont
\begin{tabular}{c|cccccccc >{\columncolor{gray!20}}c|cccccc >{\columncolor{gray!20}}c}
\hline

\hline
\multirow{2}{*}{AU} & \multicolumn{13}{c}{F1-frame} \\ \cline{2-17} 
                    & \rotatebox{90}{MLCR \cite{niu2019multi}}  & \rotatebox{90}{JAA \cite{shao2018deep}} & \rotatebox{90}{LP-Net \cite{niu2019local}} & \rotatebox{90}{ARL \cite{shao2019facial}} & \rotatebox{90}{SRERL\cite{li2019semantic}} & \rotatebox{90}{J$\rm \hat{A}$ANet \cite{shao2021jaa}}  & \rotatebox{90}{CLP \cite{li2023contrastive}} & \rotatebox{90}{MMA-Net \cite{shang2023mma}} &  \rotatebox{90}{\textbf{Ours}} & \rotatebox{90}{R-CNN* \cite{ma2019r}} & \rotatebox{90}{UGN-B* \cite{song2021uncertain}} & \rotatebox{90}{HMP-PS*\cite{song2021hybrid}} & \rotatebox{90}{DML*\cite{wang2021dual}} & \rotatebox{90}{TransAU* \cite{jacob2021facial}}  & \rotatebox{90}{ Bio-AU* \cite{cui2023biomechanics}}  & \rotatebox{90}{\textbf{Ours}}  \\ \hline
1                   & 42.4      & 47.2    & 43.3       & 45.8    & 46.9      & \textbf{53.8} & 47.7  & 52.5 & [\underline{52.6}]   & 50.2       & 54.2       & 53.1  & 52.6  & 51.7 & 57.4  & [52.6]    \\
2                   & 36.9      & 44.0    & 38.0       & 39.8    & 45.3      & {47.8} & \textbf{50.9} & \textbf{50.9} & [\underline{47.9}]  & 43.7       & 46.4       & 46.1  & 44.9  & 49.3 & 52.6  & [{47.9}]      \\
4                   & 48.1      & 54.9    & 54.2       & 55.1    & 55.6      & \underline{58.2} & 49.5 & \textbf{58.3} & [{57.3}]   & 57.0       & 56.8       & 56.0 & 56.2   & 61.0 & 64.6  & [{57.3}]    \\
6                   & 77.5     & 77.5    & 77.1       & 75.7    & 77.1      & \underline{78.5} & 75.8  & 76.3 & [\textbf{78.5}]  & 78.5       & 76.2       & 76.5  & 79.8  & 77.8 & 79.3 & [{78.5}]   \\
7                   & 77.6      & 74.6    & 76.7       & 77.2    & \underline{78.4}      & 75.8  & \textbf{78.7} & 75.7 & [{77.6}]  & 78.5       & 76.7       & 76.9  & 80.4  & 79.5 & 81.5 & [77.6]     \\
10                  & 83.6      & \underline{84.0}    & 83.8       & 82.3    & 83.5      & 82.7  & 80.2 & 83.8 & [\textbf{84.9}]  & 82.6       & 82.4       & 82.1  & 85.2  & 82.9 & 82.7 & [ \underline{84.9}]    \\
12                  & 85.8      & 86.5    & 87.2       & 86.6    & 87.6      & \underline{88.2} & 84.1  & 87.9 & [\textbf{88.4}]  & 87.0       & 86.1       & 86.4   & 88.3  & 86.3 &  85.6 & [\textbf{88.4}]      \\
14                  & 61.0    & 61.9    & 63.6       & 58.8    & {63.9}    & 63.7  & \underline{67.1} & 63.8 & [\textbf{67.8}]  & 67.7       & 64.7       & 64.8   & 65.6  & 67.6 & 67.8 & [\textbf{67.8}]      \\
15                  & 43.7   & 43.6    & 45.3       & 47.6    & \textbf{52.2}      & 43.3  & \underline{52.0} & 48.7 & [{47.6}]  & 49.1       & 51.2       & 51.5 & 51.7  & 51.9 & 47.3 & [47.6]    \\
17                  & 63.2     & 60.3    & 60.5       & 62.1    & \textbf{63.9}      & 61.8 & 62.7 & 61.7 & [\underline{63.3}]  & 62.4       & 63.1       & 63.0  & 59.4  & 63.0 & 58.0 & [\textbf{63.3}]       \\
23                  & 42.1       & 42.7    & 48.1       & \underline{47.4}    & 47.1      & 45.6  & 45.7 & 46.5 & [\textbf{47.4}]  & 50.4       & 48.5       & 49.9 & 47.3 & 43.7 & 47.0  & [47.4]     \\
24                  & 55.6      & 41.9    & {54.2}       & \textbf{55.4}    & 53.3      & 49.9 & \underline{54.8}  & 54.4 & [51.3]   & 49.3       & 53.6       & 54.5 & 49.2  & 56.3 & 44.9 & [51.3]    \\ \hline
\textbf{Avg.}                & 59.8     & 60.0    & 61.0       & 61.1    & {62.9}      & 62.4  & 62.4  & \underline{63.4}  & [\textbf{63.7}]   & 62.6       & 63.3   & 63.4 & 63.4 & 64.2 & 64.1 & [{63.7}]   \\ \hline

\hline
\end{tabular}
\end{table*}

        \subsection{Evaluation Metrics}
        For all methods, the frame-based F1 score (F1-frame, \%) is reported, which is the harmonic mean of the Precision $\rm P$ and Recall $\rm R$ and calculated by $\rm F1=2P*R/(P+R)$. To conduct a more comprehensive comparison with other methods, we also evaluate the performance with AUC (\%) refers to the area under the ROC curve and accuracy (\%). In addition, the average results over all AUs (denoted as \textbf{Avg.}) are computed with ``\%” omitted.
\begin{table}[t]
\centering
\caption{Comparisons with state-of-the-art methods for 12 AUs on BP4D in terms of Accuracy and AUC respectively (in \%). * means the method employed pretrained model on additional dataset, such as ImageNet \protect\cite{deng2009imagenet}, \textit{etc}. So we do not directly compare.} 
\vspace{-0.5em}
\label{tab:BP4D_AUC}
\setlength\tabcolsep{7pt}
\fontsize{9.5}{13}\selectfont
\begin{tabular}{c|c|ccc >{\columncolor{gray!20}} c|ccc >{\columncolor{gray!20}} c}
\hline

\hline
\multirow{2}{*}{AU}    & \multicolumn{5}{c|}{Accuracy} & \multicolumn{4}{c}{AUC} \\ \cline{2-10} 
                   & \rotatebox{90}{UGN-B* \cite{song2021uncertain}} & \rotatebox{90}{JAA \cite{shao2018deep}}  & \rotatebox{90}{ARL \cite{shao2019facial}}  & \rotatebox{90}{J$\rm \hat{A}$ANet \cite{shao2021jaa}}  & \rotatebox{90}{\textbf{MGRR-Net}} & \rotatebox{90}{DRML \cite{zhao2016deep}}   & \rotatebox{90}{SRERL \cite{li2019semantic}}  & \rotatebox{90}{DML* \cite{wang2021dual}} & \rotatebox{90}{\textbf{MGRR-Net}}  \\ \hline
1                   &78.6   & 74.7     & 73.9     & \underline{75.2}         &  \textbf{78.7}      & 55.7       & {67.6}  & \textbf{78.5}  &  \underline{78.1}       \\
2                     &80.2   & \underline{80.8}     & 76.7     & 80.2         &  \textbf{82.1}      & 54.5       & 70.0 &  \underline{75.9}    &  \textbf{77.2}       \\
4                  &80.0   & 80.4     & 80.9     & \textbf{82.9}         &  \underline{81.6}      & 58.8       & {73.4}    &  \textbf{84.4}    &  \underline{83.8}       \\
6                    &76.6  & 78.9     & 78.2     & \textbf{79.8}         &  \underline{78.7}      & 56.6   & {78.4}   &  \textbf{88.6}     &  \underline{88.4}       \\
7                     &72.3  & 71.0     & \textbf{74.4}     & 72.3         &  \underline{73.7}      & 61.0      & {76.1}  &  \textbf{84.8}    &  \underline{82.3}       \\
10                  &77.8   & \underline{80.2}     & 79.1     & 78.2  &  \textbf{81.2}      & 53.6       & {80.0}          &  \textbf{87.3}    &  \underline{86.3}       \\
12                  &84.2   & 85.4     & 85.5     & \underline{86.6}         &  \textbf{86.9}      & 60.8       & {85.9}   &  \textbf{93.9}    &  \underline{93.6}       \\
14              &63.8   & 64.8     & 62.8     & \underline{65.1}         &  \textbf{67.0}      & 57.0       & 64.4       &  \underline{71.8}    &  \textbf{72.9}       \\
15                &84.0   & 83.1     & \underline{84.7}     & 81.0         &  \textbf{84.2}      & 56.2       & 75.1    &  \underline{80.7}       &  \textbf{80.8}       \\
17                &72.8   & \underline{73.5}     & \textbf{74.1}     & 72.8         &  72.2      & 50.0       &71.7     &   \underline{75.0}     &  \textbf{78.2}       \\
23                 &82.8  & 82.3     & \underline{82.9}     & 82.9         &  \textbf{84.1}      & 53.9       & 71.6     &  \underline{78.7}      &  \textbf{79.3}       \\
24                &86.4  & 85.4     & 85.7     & \textbf{86.3}         &  \underline{86.0}      & 53.9       & 74.6     &  \underline{84.3}     &  \textbf{87.8}       \\ \hline
\textbf{Avg.}              &78.2   & 78.4     & 78.2     & \underline{78.6}         &  \textbf{79.7}      & 56.0       & 74.1  & \underline{82.0}    &  \textbf{82.4}       \\ \hline

\hline
\end{tabular}
\end{table}

    \subsection{Comparison with State-of-the-art Methods}
    We compare our proposed MGRR-Net with several frame-based AU detection baselines and the latest state-of-the-art methods, including Deep Structure Inference Network (DSIN) \cite{corneanu2018deep}, Joint AU Detection and Face Alignment (JAA) \cite{shao2018deep}, Multi-Label Co-Regularization (MLCR) \cite{niu2019multi}, Local relationship learning with Person-specific shape regularization (LP-Net) \cite{niu2019local}, Attention and Relation Learning (ARL) \cite{shao2019facial}, Semantic Relationships Embedded Representation Learning (SRERL) \cite{li2019semantic}, Joint AU detection and face alignment via Adaptive Attention Network (J$\rm \hat{A}$ANet) \cite{shao2021jaa}, Data-Aware Relation Graph Convolutional Neural network (DAR-GCN) \cite{jia2022data}, Dual-channel Graph Convolutional Neural Network (JAA-DGCN) \cite{jia2023novel}, a semi-supervised Contrastively Learning the Person-independent representations method (CLP) \cite{li2023contrastive} and a Multiview Mixed Attention based Network (MMA-Net) \cite{shang2023mma}.  
    To ensure reliable and fair comparisons, we directly use the results of these methods reported.
    Note that, the best and second-best results are shown using bold and underline, respectively. The experimental results of our MGRR-Net are shown with a grey background.
    
    For a more comprehensive display, we present methods (marked with $*$) \cite{jacob2021facial,song2021hybrid,song2021uncertain,wang2021dual,tang2021piap,chen2022geoconv, chen2022causal,cui2023biomechanics} that use additional data, such as ImageNet \cite{deng2009imagenet} and VGGFace2 \cite{cao2018vggface2}, for pre-training their complex feature extraction stem network firstly, such as ResNet \cite{he2016deep} \textit{etc}. 
    From \cite{niu2019local,jacob2021facial}, the pre-trained feature extractor improved the average F1-score by at least 1.2\% on BP4D. 
    Due to the fact that our stem network only consists of a few simple convolutional layers, even if we pre-trained on additional datasets, it is unsuitable compared to pre-training on deeper feature extraction networks, such as ResNet50 \cite{he2016deep}, ResNet101 \cite{he2016deep} and Swin Transformer-Base \cite{liu2021swin}. 
    To this end, we have grouped them together to facilitate comparison with our proposed MGRR-Net. 
    Notably, our results show excellence, affirming the superiority and efficacy of our proposed learning methodology.
    To provide a fair comparison, we omit the need for additional modality inputs and non-frame-based models \cite{liu2019multi,yang2020adaptive,shao2020spatio,yang2021exploiting,tallec2022multi}. %

    \subsubsection{Quantitative Comparison on DISFA} 
     We compare our proposed method with its counterpart in Table \ref{tab:DISFA} and Table \ref{tab:DISFA_ACC_AUC}. 
     It has been shown that our MGRR-Net outperforms all its competitors with impressive margins. 
     Compared with the existing end-to-end feature learning and multi-label classification methods DSIN \cite{corneanu2018deep} and ARL \cite{shao2019facial}, our MGRR-Net shows significant improvements on all AUs. 
     These results demonstrate the effectiveness of accurate muscle region localization for AU detection. 
     Although ARL \cite{shao2019facial} also performs sequential multiple attention explorations on global features, we believe that the sequential mechanism may destroy the diversity of different attention-aware features and slow down the training time. 
     J$\rm \hat{A}$ANet is the latest state-of-the-art method which also joint AU detection and face alignment into an end-to-end multi-label multi-branch network.  
     Compared with the baseline J$\rm \hat{A}$ANet \cite{shao2021jaa}, our MGRR-Net increases the average F1-frame and average accuracy scores by large margins of 4.7\% and 1.2\% and shows clear improvements for most annotated AU categories. 
     The main reason lies in J$\rm \hat{A}$ANet \cite{shao2021jaa} completely ignores the correlation between branches and the individual modelling of each AU. 
     Compared with JAA-DGCN \cite{jia2023novel} that also applies the graph relationship model, our MGRR-Net still performs better on most metrics because we model local relationships while supplementing a variety of information from the global face. 
     Moreover, compared with the latest state-of-the-art MMA-Net \cite{shang2023mma}, MGRR-Net achieves a 2.2\% lead in the average F1-frame metric.
     In addition, compared with the current state-of-the-art AU detection methods based on pre-trained models, such as UGN-B \cite{song2021uncertain}, HMP-PS \cite{song2021hybrid}, DML \cite{wang2021dual}, PIAP \cite{tang2021piap} and Bio-AU \cite{cui2023biomechanics} \textit{etc.}, we also achieve the best performance in terms of the average F1-frame. 
     
     Furthermore, the results of the Accuracy and AUC evaluations provide further evidence of the effectiveness of our method compared to other state-of-the-art methods. In particular, our MGRR-Net obtains a significant improvement on the average of Accuray, \textit{i.e.} 95.2 \% \textit{vs.} 94.5\%, compared with MMA-Net \cite{shang2023mma}. And on AUC metric, our MGRR-Net also achieves higher results on most metrics and increases 2.0\% compared to DML* \cite{wang2021dual}.

\begin{figure}[t] 
	\centering
	\includegraphics[width=1\linewidth]{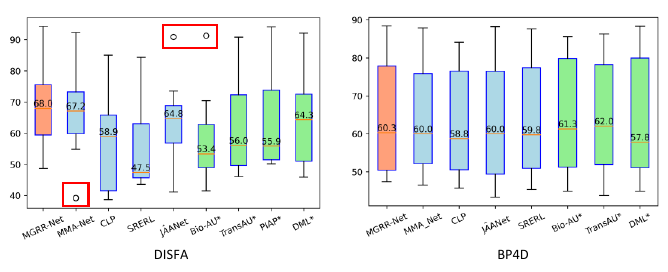}
	\vspace{-1.5em}
	\caption{Box Plots of the distribution of performances on all AU categories (the labeled values are medians). (a) on DISFA 3-flod test set and (b) on BP4D 3-flod test set. 
	}
	\label{fig:var}
	\vspace{-0.5em}
\end{figure}

    \subsubsection{Quantitative Comparison on BP4D}
    Table \ref{tab:BP4D_f1} and \ref{tab:BP4D_AUC} show the AU detection results of different methods in terms of F1-frame, Accuracy and AUC on BP4D dataset, where the method in the left of Table \ref{tab:BP4D_f1} uses a feature extractor without pre-training and the method with * is based on the pre-trained feature extractor (our method is trained on BP4D only). 
    Compared with the multi-branch combination-based J$\rm \hat{A}$ANet \cite{shao2021jaa}, the average F1 frame score and average accuracy score of MGRR-Net get 1.3\% and 1.1\% higher, respectively. 
    Furthermore, compared with the latest graph-based relational modelling method SRERL \cite{li2019semantic}, MGRR-Net increases the average F1-frame and average AUC by large margins of 0.8\% and 8.3\%. 
    This is mainly due to the fact that the proposed method models the semantic relationships among AUs while also gaining complementary features from multiple global perspectives to increase the distinguishability of each AU.   
    In addition, our MGRR-Net achieves the best or second-best AU detection performance in terms of F1-frame, Accuracy and AUC for most of the 12 AUs annotated in BP4D compared with the state-of-the-art methods. 
    For example, compared with the latest method MMA-Net \cite{shang2023mma}, which simultaneously modelled the deep feature learning and the structured AU relationship in a unified framework, ours greatly outperforms it by 0.3\% in terms of the average of F1-frame. 
    In addition, compared with the advanced models pre-trained with additional data (marked with $*$ in Table \ref{tab:BP4D_f1} and Table \ref{tab:BP4D_AUC}), our MGRR-Net still has strong competitiveness. 
    
    Experimental results of MGRR-Net demonstrate its effectiveness in improving AU detection accuracy on DISFA and BP4D, as well as good robustness and generalization ability. 
    Note that the main reason why some AUs are clearly less accurate than others is due to data imbalance, as shown in Figure \ref{fig:var}, this is a phenomenon that exists in all existing methods \cite{shang2023mma,tang2021piap,cui2023biomechanics,shao2021jaa,li2023contrastive,jacob2021facial}. In BP4D, where the data distribution is relatively reasonable, the results' distribution of each method is close. But in DISFA, where the data distribution is more extreme, the result distribution of our MGRR-Net can perform better, \textit{i.e.} lower variance and no outliers. We infer that two aspects promote this improvement. On one hand, we use a weighted multi-label cross-entropy loss function as Eq.(\ref{wCE}) to solve the data imbalance problem to a certain extent. On the other hand, our multi-level fused representation can complement each AU representation, as well as combine with other AU areas, to further improve AU classification.

\begin{table*}[t]
\centering
\fontsize{9.5}{13}\selectfont
\renewcommand\tabcolsep{7.0pt}
\caption{Effectiveness of key components of MGRR-Net evaluated on DISFA in terms of F1-frame score (in \%).} 
\label{tab:Ablation_DISFA}
\begin{tabular}{cc|cccccc >{\columncolor{gray!20}} c}
\hline
\multicolumn{2}{c|}{Method}                           & 1 & 2 & 3 & 4 & 5 & 6 & MGRR-Net \\ \hline
\multicolumn{1}{c|}{\multirow{4}{*}{\rotatebox{90}{Setting}}}  & D\_G & -  & $\surd$  & $\surd$  & $\surd$   &$\surd$  & $\surd$ & $\surd$  \\
\multicolumn{1}{c|}{}                          & O\_G & -                    & -  & $\surd$  & -  & $\surd$ & $\surd$ & $\surd$ \\
\multicolumn{1}{c|}{}                          & C\_G & -                    & -  &  -     & $\surd$  & $\surd$  & -  & $\surd$\\
\multicolumn{1}{c|}{}                          & P\_G & -                    & -  &  -     & $\surd$  & -  & $\surd$ & $\surd$ \\ \hline
\multicolumn{1}{c|}{\multirow{8}{*}{\rotatebox{90}{AU Index}}} & 1 & 47.1  & 52.5  & 58.4  & 60.0  &  65.4 & 61.0  & [61.3]\\
\multicolumn{1}{c|}{}                          & 2    & 61.1  & 58.1  & 63.0  & 65.7  &  64.5 & 67.3  &[62.9]\\
\multicolumn{1}{c|}{}                          & 4    & 66.3  & 73.3  & 70.9  & 67.4  &  72.5 & 76.8  & [75.8] \\
\multicolumn{1}{c|}{}                          & 6    & 44.7  & 44.4  & 46.2  & 43.8  &  42.6 & 40.9  & [48.7]\\
\multicolumn{1}{c|}{}                          & 9    & 52.2  & 52.5  & 47.7  & 57.1  &  52.9 & 58.0  & [53.8]\\
\multicolumn{1}{c|}{}                          & 12   & 74.9  & 73.2  & 72.1  & 75.4  &  75.3 & 74.8  & [75.5]\\
\multicolumn{1}{c|}{}                          & 25   & 92.2  & 94.7  & 93.4  & 93.3  &  94.3 & 93.7  & [94.3]\\
\multicolumn{1}{c|}{}                          & 26   & 66.2  & 71.2  & 71.8  & 64.7  &  71.4 & 65.8  & [73.1]\\ \hline
\multicolumn{2}{c|}{Avg.}                             & 63.1  & 65.0  & 65.4  & 65.9  &  67.4 & 67.3  & [68.2]\\ \hline
\end{tabular}
\end{table*}
    \subsection{Ablation Studies}
    We perform detailed ablation studies on DISFA to investigate the effectiveness of each part of our proposed MGRR-Net. 
    Due to space limitations, we do not show the ablation results for BP4D, but it is consistent with DISFA.
    To assess the effect of different components, we run the experiments with same parameter setting (\textit{e.g.} layer K=2) for variations of the proposed network in Table \ref{tab:Ablation_DISFA}.

    \subsubsection{Effects of Region-level Dynamic Graph}
    In Table \ref{tab:Ablation_DISFA}, we can see that learning by the dynamic graph initialized with prior knowledge (indicated by D\_G) outperforms baseline with an improvement of average F1-frame from 63.1\% to 65.0\%, indicating that the dynamic graph could get richer features from other correlated AU regions to improve robustness. 
    Furthermore, to cancel out the initialization of prior knowledge, we randomly initialize the dynamic graph, which decreases F1-frame to 64.7\%. 
    These observations suggest that the relationship reasoning in the dynamic graph can significantly boost the performance of AU detection, while prior knowledge makes a great contribution but not predominantly.

    \subsubsection{Effects of Multi-level Global Features} 
    We test the contributions of multiple important global feature components of the model in Table \ref{tab:Ablation_DISFA}, namely, original global feature (O\_G) from stem network, channel-level global feature (C\_G) from channel-level MH-GAT and pixel-level global feature (P\_G) from pixel-level MH-GAT.  
    After we supplemented original global feature (O\_G) for each target AU, the average F1-frame score has been improved from 65.0\% to 65.4\%, demonstrating the effectiveness of global detail supplementation. 
    The fusion of channel- and pixel-level global features (C\_G and P\_G) results in a 0.9\% increase, indicating that they make the AU more discriminative than only using the original global features. 
    Comparing the results of the fifth test (with C\_G) and the sixth test (with P\_G) in Table \ref{tab:Ablation_DISFA} with the third test, one of the channel-level and pixel-level global features can boost the performance by roughly the same amount. 
    It suggests that by supplementing and training different levels of global features for each AU branch, more global details can be provided to detect AUs in terms of different expressions and individuals. 
    
    Finally, the hierarchical gated fusion of multi-level global and local features leads to a significant performance improvement to 68.2\% in terms of F1-frame score. 
    It validates that the dynamic relationship of multiple related face regions provides more robustness, while the supplementation of multi-level global features makes the AU more discriminative. 

\begin{table}[t]
\centering
\fontsize{9.5}{12.5}\selectfont
\renewcommand\tabcolsep{3pt}
\caption{Performance comparison of MGRR-Net with different iteration step number $\rm K$ on DISFA in terms of F1-frame score (in \%).} 
\label{tab:layers}
\begin{tabular}{cccccccccc}
\hline

\hline
\multirow{2}{*}{Layers} & \multicolumn{8}{c}{AU Index}                          & \multirow{2}{*}{\textbf{Avg.}} \\ \cline{2-9}
                        & 1 & 2 & 4 & 6 & 9 & 12 & 25 & 26 &                       \\ \hline
K=1  & \underline{64.5}  & 58.3  & 74.9  & 46.1  & \textbf{54.4}  &  75.4  &  92.3  & 73.1  & \underline{67.4} \\
\rowcolor{gray!20}  K=2  & 61.3  & \underline{62.9}  & \underline{75.8}  & \textbf{48.7}  & \underline{53.8}  & \textbf{75.5}   & \textbf{94.3}   & \textbf{73.1}  & \textbf{68.2}  \\
K=3  & \textbf{65.5}  & \textbf{67.0}  & \textbf{77.6}  & 40.0  & 44.9  & 75.1   & 94.0   & 68.8   & 66.6 \\ \hline

\hline
\end{tabular}
\vspace{-0.5em}
\end{table}    

\begin{table}[t]
\centering
\fontsize{9.5}{12.5}\selectfont
\renewcommand\tabcolsep{4pt}
\caption{Mean error (\%) results of different face alignment models on DISFA and BP4D (lower is better). }
\label{tab:landmark}
\begin{tabular}{c|ccc >{\columncolor{gray!20}} c}
\hline

\hline
Datasets & MCL    & JAA      & J$\rm \hat{A}$ANet & \textbf{MGRR-Net} \\ \hline
DISFA   & 7.15  & 6.30 &  4.02      & \textbf{3.95}    \\ 
BP4D    & 7.20     & 6.38        &  \textbf{3.80}      &  4.01     \\ 
\hline

\hline
\end{tabular}
\vspace{-0.5em}
\end{table}    
    \subsubsection{Effects of Layer Number}
    We evaluate the impact of layer number of our proposed iterative reasoning network. 
    As shown in Table \ref{tab:layers}, MGRR-Net achieves the averaged F1-frame score of 67.4\%, 68.2\% and 66.6\% on DISFA when the reasoning layer number K is set to 1, 2 and 3 respectively. 
    The averaged F1-frame scores on BP4D dataset are 63.5\%, 63.7\%, and 63.1\% respectively.  
    It achieves the best performance when K=2, and is overfitted when K$>$2.
    Finally, the optimal number of layers is 2 for our MGRR-Net on DISFA and BP4D datasets.

\begin{figure}[t] 
	\centering
	\vspace{-0.5em}
	\includegraphics[width=0.8\linewidth]{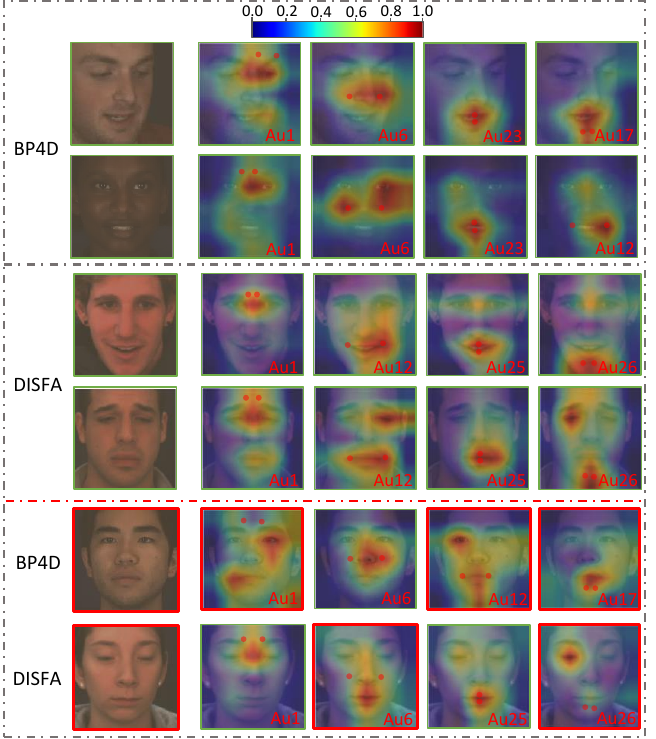}
	\caption{Class activation maps that show the discriminative regions for different AUs in terms of different expressions and individuals on DISFA and BP4D datasets. We show the region center positions defined by the detected landmarks for the corresponding AUs. Abnormally shifted AU activation maps are marked with red boxes.
	}
	\label{fig:visual_01}
\end{figure}  
    \subsubsection{Results for Face Alignment}
    We jointly take face alignment network into our MGRR-Net via auxiliary training, which can provide effective muscle regions corresponding to AUs based on the detected landmarks. 
    Table \ref{tab:landmark} shows the mean error results of our MGRR-Net and baseline method J$\rm \hat{A}$ANet \cite{shao2021jaa} on DISFA and BP4D. 
    We also compare with state-of-the-art face alignment methods that have released trained models, including MCL \cite{shao2020deep}, JAA \cite{shao2018deep}. 
    Our MGRR-Net achieves competitive 3.95 and 4.01 mean errors on DISFA and BP4D respectively. 
    It indicates that with the comparable face alignment performance as J$\rm \hat{A}$ANet, our MGRR-Net can achieve better AU detection accuracy.

    \subsection{Visualization of Results}
\begin{figure}[t] 
	\centering
	\includegraphics[width=0.8\linewidth]{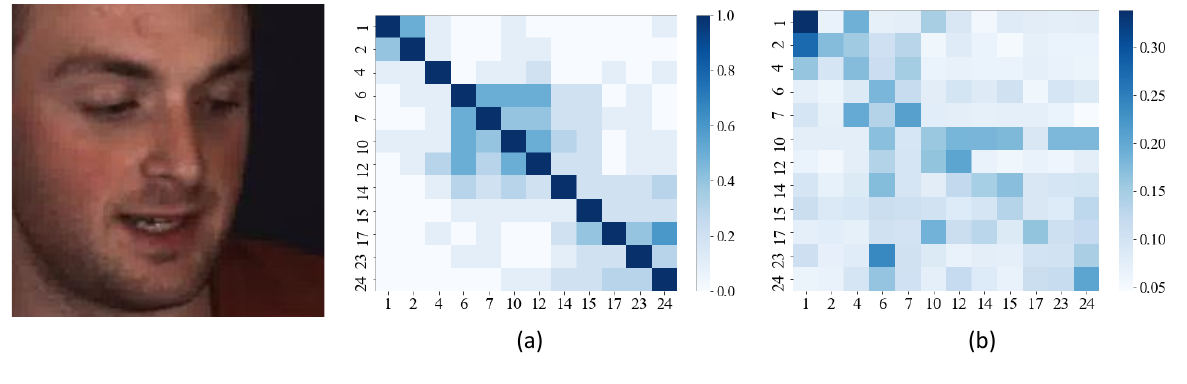}
	\vspace{-1.5em}
	\caption{Visualizations of the predefined AU correlation (a) and the learned relevance matrix (b) for the individual on BP4D.  The corresponding class activation maps are shown in the first row of Figure \protect\ref{fig:visual_01}. 
	}
	\label{fig:visual_02}
	\vspace{-0.5em}
\end{figure}

    To better understand the effectiveness of our proposed model, we visualize the learned class activation maps of MGRR-Net corresponding to different AUs in terms of different expressions, postures and individuals, as shown in Fig. \ref{fig:visual_01}. 
    Three examples are from DISFA and three are from BP4D (Two bad examples of abnormal offsets happening are shown at the bottom of Fig. \ref{fig:visual_01}.), containing visualization results of different genders and different poses with different AU categories. 
    Through the learning of MGRR-Net, not only the concerned AU regions can be accurately located, but also the positive correlation with other AU areas can be established and other details of the global face can be supplemented. 
    The different activation maps of the same AU on different individuals show that our MGRR-Net can dynamically adjust according to the differences of expression, posture, and individual. 
    Some activation maps are inconsistent with the predefined AU areas, which may be caused by the insensitivity to the target predefined areas after the introduction of multi-level global supplementation. 
    In addition, as shown in Fig. \ref{fig:visual_02}, we further visualize the learned relevance matrix (marked as (b)) and the predefined AU correlations (marked as (a)) of the individual corresponding to the first row of Fig. \ref{fig:visual_01} on BP4D. 
    The predefined correlation matrices are used to roughly calculate the co-occurrence relevance between different AUs by counting the dependence of positive and negative samples. 
    It over-emphasises target AU as well as a few other AUs while other AU regions are completely ignored, due to bias in the statistics of the data. 
    From the correlation matrix we learned, the target AU and the relevant AU are highlighted without discarding information from other branches at all, which is beneficial for increasing the distinguishability between AUs. 
    Furthermore, the supplementation of global features with multiple perspectives allows different AUs to access a lot of information outside the defined areas, as shown in Fig. \ref{fig:visual_01}, which is helpful for adaptive changes in terms of different individuals and their expressions. 

\section{Conclusion}
    In this paper, we have proposed a novel multi-level graph relational reasoning network (termed MGRR-Net) for facial AU detection. 
    Each layer of MGRR-Net can encode the dynamic relationships among AUs via a region-level relationship graph and multiple complementary levels of global information covering expression and subject diversities.
    The multi-layer iterative feature refinement finally obtains robust and discriminative features for each AU. 
    Extensive experimental evaluations on DISFA and BP4D show that our MGRR-Net outperforms state-of-the-art AU detection methods with impressive margins. 
    
    In our future work, we will introduce the pre-trained models to improve the performance of the stem network in extracting feature representation, and we would like to investigate the implementation of facial AU detection into real applications, such as automatically estimating facial palsy severity for patients. 
    This will be helpful for the diagnosis and treatment of people who have facial palsy across the world. 
    In collaboration with medical professionals, we will collect and annotate facial palsy datasets, such as \cite{o2010objective}, to further validate the migration capability and effectiveness of the proposed model.


\begin{acks}
This research has been supported in part by the National Natural Science Foundation of China (No. 62176249) and in part by the China Scholarship Council (CSC) from the Ministry of Education of China (No.202006310028).
\end{acks}

\bibliographystyle{ACM-Reference-Format}
\bibliography{sample-base}


\end{document}